\documentclass[a4paper]{cas-sc}

\usepackage[numbers]{natbib}
\def\tsc#1{\csdef{#1}{\textsc{\lowercase{#1}}\xspace}}
\tsc{WGM}
\tsc{QE}
\renewcommand{\paragraph}[1]{%
  \vspace{6pt}\noindent\textbf{#1}\enspace}

\begin{document}
\let\WriteBookmarks\relax
\def\floatpagepagefraction{1}
\def\textpagefraction{.001}

\shorttitle{Graph Machine Learning: An Opportunity for Power Systems}    

\shortauthors{M. Sadric et al.}

\title [mode = title]{Graph Machine Learning: An Opportunity for Power Systems}  

\tnotemark[1]
\tnotetext[1]{This work is funded by the Helmholtz Association's Initiative 
and Networking Fund through Helmholtz AI, the Helmholtz Association under grant 
no. VH-NG-1727, and co-funded by the Federal Ministry for Economic Affairs and 
Energy (BMWE), project FKZ 03EI1092C.}



\author[1]{Martin Sadric}[orcid={0009-0000-5678-0390}]
\fnmark[1]
\ead{martin.sadric@kit.edu}
\credit{Conceptualization, Methodology, Formal analysis, Investigation, Data curation, Writing -- original draft, Visualization}

\author[1,2]{Sebastian Pütz}[orcid={0009-0009-8468-4166}]
\fnmark[1]
\ead{sebastian.puetz@kit.edu}
\credit{Conceptualization, Methodology, Investigation, Writing -- original draft, Supervision, Project administration}

\author[3]{Christian Nauck}[orcid=0000-0003-1972-9654]
\ead{christian.nauck@pik-potsdam.de}
\credit{Writing -- review and editing}

\author[1]{Veit Hagenmeyer}[orcid=0000-0002-3572-9083]
\ead{veit.hagenmeyer@kit.edu}
\credit{Writing -- review and editing}

\author[3]{Frank Hellmann}[orcid=0000-0001-5635-4949]
\ead{frank.hellmann@pik-potsdam.de}
\credit{Writing -- review and editing}

\author[4]{Dirk Witthaut}[orcid=0000-0002-3623-5341]
\ead{d.witthaut@fz-juelich.de}
\credit{Writing -- review and editing}

\author[1,2]{Benjamin Schäfer}[orcid=0000-0003-1607-9748]
\cormark[1]
\ead{benjamin.schaefer@kit.edu}
\credit{Writing -- review and editing, Supervision, Project administration, Funding acquisition}

\affiliation[1]{organization={Karlsruhe Institute of Technology},
                addressline={Kaiserstraße 12},
                city={Karlsruhe},
                postcode={76131},
                country={Germany}}

\affiliation[2]{organization={Helmholtz AI},
                country={Germany}}

\affiliation[3]{organization={Potsdam Institute for Climate Impact Research},
                addressline={Telegrafenberg A 31},
                city={Potsdam},
                postcode={14473},
                country={Germany}}

\affiliation[4]{organization={Forschungszentrum Jülich},
                addressline={Wilhelm-Johnen-Straße},
                city={Jülich},
                postcode={52428},
                country={Germany}}

\cortext[cor1]{Corresponding author}
\fntext[fn1]{These authors contributed equally}
\begin{abstract}
Modern power systems face growing operational complexity driven by the integration of renewable energy sources, decentralization, and the need for real-time decision-making across a wide range of timescales. Addressing these challenges traditionally relies on model-based methods that, while accurate, can be too slow for operational demands. Machine learning (ML) has therefore emerged as a faster, data-driven alternative. As grid topology plays a central role in power system operation, graph machine learning (GML) methods offer a natural framework for incorporating topological dependencies as an inductive bias. We survey nearly 800 papers at the intersection of GML and power systems, covering forecasting, state estimation, optimization, control, fault diagnosis, and cybersecurity. Power systems constitute an unusually rich benchmark setting for GML, as they combine hard physical constraints, multi-scale dynamics, safety-critical requirements, and scarce labeled data within a single, well-defined domain. Conversely, power systems can benefit from utilizing GML to complement classical solvers, as GML provide scalable, topology-aware approximations with promising generalization and computational efficiency. We identify open challenges, including limited real-world deployment and the need for interpretable models in safety-critical settings. Despite the rapidly growing number of publications, standardized benchmarks and open datasets remain scarce, leaving many results difficult to reproduce and undermining the long-term scientific credibility of the field. We further derive a structured requirements catalog for ML-ready power grid benchmarks, intended to guide future dataset development and improve reproducibility across studies. We call on the community to prioritize dedicated benchmark studies and the release of open datasets and models. 
\end{abstract}

\begin{graphicalabstract}
    \centering
    \includegraphics[width=\linewidth]{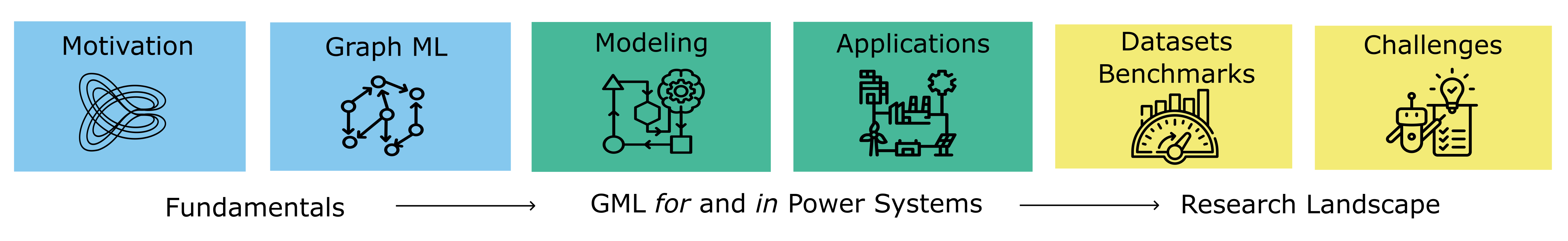}

    \small\textbf{Structure of the article}. The increasing complexity of modern power systems calls for machine learning methods that are natively aware of the underlying graph structure. This review traces that argument from \textbf{motivation} through \textbf{Graph ML} (GML) fundamentals and \textbf{modeling} of power system physics as graphs, to a survey of GML \textbf{applications} across forecasting, optimization, control, and reliability. We close by assessing available \textbf{datasets and benchmarks} and identifying open \textbf{challenges} that limit real-world deployment.
\end{graphicalabstract}

\begin{highlights}
\item Systematic survey of nearly 800 papers on graph machine learning for power systems
\item Graph ML encodes grid topology as inductive bias for scalable power system tasks
\item Applications span forecasting, state estimation, optimization, fault diagnosis, cybersecurity
\item Open datasets and benchmarks remain scarce, limiting reproducibility in the field
\item Closer collaboration between ML and power systems communities is needed
\end{highlights}

% Keywords
\begin{keywords}
graph machine learning \sep  graph neural networks \sep power systems \sep energy systems

\end{keywords}

\maketitle

% Main text
% 1. Introduction
% Goal: Hook the reader immediately and establish the necessity of this intersection.

% What is it about?
% The growing complexity of power grids (renewables, decentralization) requires faster, scalable computation.
% Why GML? The grid is a graph. This is the natural mathematical language for power systems.
% How this review was written: Briefly mention the methodology, make a reference to the webpage (no appendix)

% What is it not about?
% It is not a generic review of AI in energy (like RL or standard NN). It is specifically about Graph structures.

% Audience Goals:
% PSE: "Why should I care? My physics-based solvers work fine." -> Answer: Speed and scalability for real-time operations.
% ML: "Is this just another dataset?" -> Answer: No, this is a domain with strict physical constraints (Kirchhoff’s laws) that challenges standard GNN architectures.

% general motivation
\section{Introduction}
The urgent need to mitigate climate change is accelerating the transition toward sustainable and carbon-neutral energy systems~\cite{rogeljEnergySystemTransformations2015}. 
% This transformation poses significant challenges for modern power systems, including the decentralization of generation~\cite{ackermannDistributedGenerationDefinition2001}, reduced system inertia~\cite{milanoFoundationsChallengesLowInertia2018}, the necessity for grid expansions~\cite{titz2024identifying}, and the shift from synchronous machines to inverter-based resources~\cite{milanoFoundationsChallengesLowInertia2018}.
% 
% Vorschlag für weniger Dopplung
This transformation poses significant challenges for modern power systems, including the decentralization of generation~\cite{kabeyiSustainableEnergyTransition2022}, the necessity for grid expansions~\cite{titzIdentifyingDriversMitigators2024a}, and reduced system inertia with the shift from synchronous machines to inverter-based resources~\cite{milanoFoundationsChallengesLowInertia2018}.
At the same time, power systems are critical infrastructure where failures can have severe societal and economic consequences.
Machine Learning (ML) can help to address these challenges and thus contribute to a secure and sustainable energy supply~\cite{ibrahimMachineLearningDriven2020a, khodayarDeepNeuralNetworks2023}. Physics-based models and optimization methods often struggle to scale to the complexity and size of modern power systems within the time constraints of real-time operation. Beyond computational tractability, accurate modeling is increasingly hindered by limited observability, particularly at the distribution grid level where measurement infrastructure remains sparse, posing significant challenges for classical simulation and state estimation approaches. ML offers a data-driven complement to these methods, capable of learning complex system behavior from historical data and delivering forecasts, control policies, and situational awareness at operational timescales.

% introduce gml
While ML provides scalable, data-driven tools applicable to the power system domain, many methods fail to capture the relational, structured, and spatial characteristics inherent in power system data. Graph Machine Learning (GML) differs from conventional ML by operating directly on entire graphs rather than fixed-size feature vectors and is thus particularly well-suited to power systems. GML has proven effective across domains as diverse as social networks, chemistry, and physics.  In the present review, we focus specifically on electrical power systems, which encompass the generation, transmission, distribution, and consumption of electricity.
Accordingly, our discussion centers on the physical power grid and its components, most prominently buses (generators or consumers), transmission lines, and transformers, which can be naturally represented as graphs (see Section~\ref{s:power_grid_is_a_graph}).
This graph-native character is shared with other infrastructure networks, such as gas and transportation, enabling direct methodological exchange in both directions. A further key advantage is that GML models can be inductive. Once trained, they have the potential to generalize to other graphs with different topology, enabling the application of the same model across power grids of varying size and structure~\cite{nauckDynamicStabilityAssessment2023}.

% one to rule them all
% Whether GML is suitable for a given power system application depends heavily on the problem context. Power systems operate across an extreme range of timescales~\cite{machowskiPowerSystemDynamics2020}: wave and electromagnetic transients (picoseconds to milliseconds, covering switching surges, fault protection, and inverter dynamics), electromechanical dynamics (seconds to minutes, covering frequency and voltage regulation and transient stability), thermodynamic dynamics (minutes to hours, covering economic dispatch and scheduling), and long-term planning (years, covering network expansion and investment). For GML purposes, these collapse into three methodologically distinct regimes: fast dynamics governed by differential-algebraic equations requiring real-time control; medium-timescale operational optimization (e.g. OPF, unit commitment); and long-term planning problems formulated as large-scale investment optimization. No single GML approach spans all regimes, and methodological choices must be made accordingly.

% shortly remember this is a review
\paragraph{Contributions.}
The rapid growth of GML publications in power systems makes a systematic review timely. Building on and extending the survey by~\citet{liaoReviewGraphNeural2022}, we incorporate research trends that have emerged since 2021, highlighting new application areas and place a stronger emphasis on interdisciplinary perspectives. We survey articles published until the end of 2025, map them to core power system tasks, and identify critical technical and methodological gaps. This review targets two primary audiences: For power system engineers and researchers, we explain where and why GML represents a leading approach; for ML and GML researchers, we motivate the development of new methods and reproducible benchmarks for a particularly relevant application. A recurring finding of our survey is the scarcity of standardized benchmarks and open datasets and the need for methodological rigor.
%: as the number of GML publications grows, results that cannot be reproduced or verified accumulate, weakening the scientific basis of the field. We advocate throughout for methodological rigor: as the field matures, the ability to reproduce and compare results across studies becomes as important as predictive accuracy itself.

% We provide a systematic literature survey of GML usage for power systems including articles until end of 2025.

\paragraph{Structure}
% paper organization 
This review is organized as follows:  Section~\ref{s:graph_machine_learning} establishes the GML foundations required throughout the paper. Section~\ref{s:power_grid_is_a_graph} details how power systems are translated into graph representations. Having established these foundations, we present our systematic survey of GML applications (Section~\ref{s:applications}) and assess the current benchmark landscape (Section~\ref{s:datasets_benchmarks}), including a requirements catalog for ML-ready power grid datasets. Finally, we identify numerous future research opportunities in Section~\ref{s:challenges_and_outlook}. 
%A visual overview of this structure is provided in Fig.~\ref{fig:graphical_abstract}. 
Supplementary materials such as reference databases, including all identified articles, evaluation metrics, and detailed annotations are provided on our companion web page: \url{https://kit-iai-dracos.github.io/graph-ml-power-sys/}.

\section{Graph Machine Learning}
\label{s:graph_machine_learning}

\begin{figure*}
    \centering
    \includegraphics[width=\linewidth]{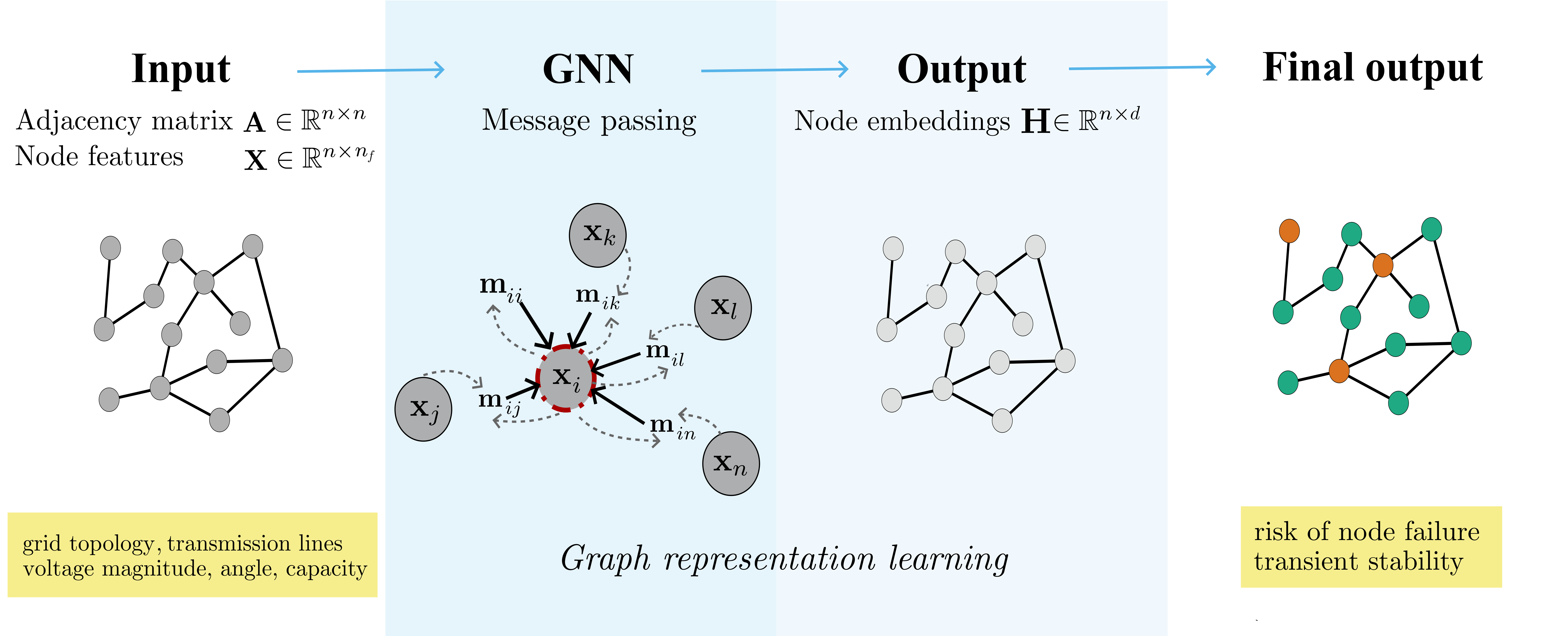}
  \caption{
{GNN workflow for power grid tasks.}
The model takes as input the adjacency matrix $\mathbf{A} \in \mathbb{R}^{n \times n}$ 
encoding grid topology and node feature matrix $\mathbf{X} \in \mathbb{R}^{n \times n_f}$ 
(e.g., voltage magnitude, angle, capacity).
Through \textit{graph representation learning}, the GNN employs \textit{message passing} 
to iteratively aggregate neighborhood information, mapping the raw graph-structured input 
into node embeddings $\mathbf{H} \in \mathbb{R}^{n \times d}$.
These are then used for predictions at the node level 
($\hat{\mathbf{Y}} \in \mathbb{R}^n$, e.g., risk of node failure) 
or at the graph level ($\hat{Y} \in \mathbb{R}$, e.g., gird transient stability assessment).
}

    \label{fig:gnn_workflow}
\end{figure*}

% general GML definition and context
Graph machine learning (GML) is a branch of machine learning concerned with modeling and learning from graph-structured data, encompassing graph kernels, spectral methods, and Graph Neural Networks (GNNs). Throughout this review, we use the term GML rather than GNNs to reflect this broader scope, even though Message Passing Neural Networks (MPNNs) currently dominate the power systems literature.

% gml teaser - wichtig: from "low-dimensionality and structure of the physical world" to
~\citet{bronsteinGeometricDeepLearning2021} provide a deep mathematical foundation for geometric deep learning architectures, GML being one of them. They argue that, although learning generic functions in high dimensions poses a challenging estimation problem, most relevant tasks are not generic. Rather, they possess essential predefined regularities stemming from the underlying structure of the physical world. The authors identify graphs as one of the five fundamental structures. Graphs have no canonical node ordering, i.e., the same graph can be represented by many equivalent adjacency matrices that are related by permutations. Graphs are also a generalization of some other structures, such as grids and sets. This means that many other deep learning architectures (e.g. Convolutional Neural Networks (CNNs) and transformers) can be expressed in the language of GML.
% history of GNNs (early models + spectral GNNs)
GNNs were first introduced in the early 2000s as recurrent models designed to handle structured data~\cite{goriNewModelLearning2005}. The field advanced significantly with the introduction of spectral GNNs inspired by graph signal processing~\cite{brunaSpectralNetworksLocally2014a, defferrardConvolutionalNeuralNetworks2016a, kipfSemiSupervisedClassificationGraph2016}, discussed further below.

In the following, we first lay general graph foundations and explain how GML uses message passing and how GML works beyond message passing. 

\subsection{Foundations}
% message-passing formalism and dominance
\paragraph{Message Passing GNNs.}
A major milestone in GML came with the work by~\citet{pmlr-v70-gilmer17a}, which unified many earlier architectures~\cite{brunaSpectralNetworksLocally2014a, defferrardConvolutionalNeuralNetworks2016a, kipfSemiSupervisedClassificationGraph2016} under the common framework of message passing, defining GNNs as models that iteratively update node representations by exchanging information between neighboring nodes (Fig.~\ref{fig:gnn_workflow}). MPNNs remain the most widely used approach in modern graph learning due to their flexibility, efficiency, and effectiveness across domains.

\paragraph{Graph Definitions.}
A graph is defined as $\mathcal{G} = (\mathcal{V}, \mathcal{E})$, where $\mathcal{V}$ is the set of $n$ nodes and $\mathcal{E} \subseteq \mathcal{V} \times \mathcal{V}$ is the set of edges. Nodes and edges may carry feature vectors, denoted $\mathbf{x}_v$ and $\mathbf{x}_{vu}$ respectively. Grid topology is encoded in the adjacency matrix $\mathbf{A}$ and graph Laplacian $\mathbf{L} = \mathbf{D} - \mathbf{A}$:
\begin{align}
    A_{ij} &= \begin{cases} 1 & \text{if } (i,j) \in \mathcal{E} \\ 0 & \text{otherwise} \end{cases}, 
    \qquad D_{ii} = \sum_j A_{ij}.
\end{align}
Since $\mathbf{A}$ depends on the choice of node ordering, permuting nodes by $\mathbf{P}$ yields an equivalent representation $\mathbf{P}\mathbf{A}\mathbf{P}^\top$, motivating the requirement that GNNs be \textit{permutation-equivariant}.

\paragraph{Node Features and Neighborhoods.}
The nodal feature matrix $\mathbf{X} \in \mathbb{R}^{n \times n_f}$, where $n_f$ is the number of input features, and the adjacency matrix $\mathbf{A}$ together constitute the two main inputs to a GNN (Fig.~\ref{fig:gnn_workflow}). More advanced architectures may additionally include edge features and global graph-level features~\cite{hamiltonGraphRepresentationLearning2020}. Let $\mathcal{N}_v$ denote the set of nodes adjacent to $v$. A single message-passing step exchanges information between directly connected nodes.

\paragraph{Message Passing Formalism.}
A node embedding $\mathbf{h}_v^{(l)}$ is a learned vector representation encoding both node features and graph context, analogous to a compact reduced-order state in physical modeling. Starting from $\mathbf{h}_v^{(0)} = \mathbf{x}_v$, embeddings are updated layer by layer. The general update rule is:
\begin{equation}
\label{eq:mpnn_general}
\mathbf{m}_{uv}^{(l)} = \phi \left( \mathbf{h}_v^{(l)}, \mathbf{h}_u^{(l)}\right), \qquad
\mathbf{h}_v^{(l+1)} = \psi \left( \mathbf{h}_v^{(l)}, \bigoplus_{u \in \mathcal{N}_v} \mathbf{m}_{uv}^{(l)} \right),
\end{equation}
where $\mathbf{m}_{uv}^{(l)}$ is the message from neighbor $u$ to node $v$, $\bigoplus$ is a permutation-invariant aggregation (e.g., sum, mean, or max), and $\phi$, $\psi$ are learnable functions. A simple instantiation averages neighbor embeddings directly:
\begin{equation}
\mathbf{h}_v^{(l+1)} = \sigma\left( \mathbf{W}^{(l)} \cdot \frac{1}{|\mathcal{N}_v|} \sum_{u \in \mathcal{N}_v} \mathbf{h}_u^{(l)} + \mathbf{B}^{(l)} \cdot \mathbf{h}_v^{(l)} \right),
\label{eq:mpnn_example}
\end{equation}
where $\mathbf{W}^{(l)}$ and $\mathbf{B}^{(l)}$ are learnable weight matrices and $\sigma$ a nonlinear activation, such as ReLU. The final embedding after $L$ layers, $\mathbf{z}_v = \mathbf{h}_v^{(L)}$, is used for downstream prediction. The choice of $\phi$ defines the architecture, with many variants available in unified frameworks such as PyTorch Geometric~\cite{Fey/Lenssen/2019, feyPyG20Scalable2025}. 

\paragraph{Graph Tasks.}
The message-passing procedure from equ.~\eqref{eq:mpnn_general} is permutation-equivariant, i.e., the final representation of a node depends only on its own features and those of its neighbors, but not on the order in which the neighbors are considered. The node embeddings from the final layer, $\mathbf{h}_v^{(L)}$, can be collected into a matrix $\mathbf{H}$, and used to produce a task-specific output $\mathbf{\hat{Y}}$ at the node, graph, or edge level (Fig.~\ref{fig:gnn_workflow}). Node- and edge-level outputs are obtained by directly transforming the corresponding embeddings. Graph-level outputs additionally require a readout (pooling) step that aggregates all node embeddings. This can be realized by simple permutation-invariant operations such as sum, mean, or max, or by a learned pooling network.

\subsection{MPNNs for Power Systems}
Power system graphs differ from the graphs for which standard MPNNs were originally 
designed in several important ways. Nodes and edges are not homogeneous entities: 
Buses, generators, loads, and branches are physically distinct components governed 
by different equations and carrying different parameters. Edge features such as 
impedance, admittance, and power flow are not merely auxiliary attributes but the 
primary carriers of physical meaning, are inherently directional, and the relevant 
prediction target is often a branch (edge) quantity rather than a nodal one. Standard GNN 
formulations therefore require several adaptations to fully respect this structure.

\paragraph{Homophily and Heterophily.}
Homophily, the tendency of nodes to connect to similar 
nodes~\cite{newmanNetworksIntroduction2016}, is a 
common assumption in many GNN architectures. In power systems, both cases arise: 
Neighboring nodes share similar properties in weather-driven forecasting 
applications~\cite{karimiSpatiotemporalGraphNeural2021, linSpatialTemporalResidentialShortTerm2021}, but the diversity of grid components 
and their interactions makes heterophilic modeling often more appropriate. In the power flow setting for instance, homogeneous GNN approaches have been shown to inadequately represent the varied node and edge types present in real-world grid topologies \cite{ghamiziOPFHGNNGeneralizableHeterogeneous2024}. Heterogeneous graphs assign distinct types to nodes and edges~\cite{hamiltonGraphRepresentationLearning2020}, allowing buses, generators, loads, and branches to maintain separate feature spaces and type-specific aggregation schemes. 
Furthermore, hypergraphs generalize edges to connect more than two nodes simultaneously, enabling representation of multi-bus constraints or control areas, and have been applied to power flow problems~\cite{gaoHGNNGeneralHypergraph2023}. 

\paragraph{Edge Features and Line Graph.}
Edge features are incorporated by conditioning the message function on edge 
attributes:
\begin{equation}
\phi\left(\mathbf{h}_u^{(l)}, \mathbf{h}_v^{(l)}, \mathbf{e}_{uv}\right),
\end{equation}
yielding weighted aggregation where edges with different physical properties 
produce different messages. Since the flow of power is directional, each line is represented as 
two directed half-edges carrying distinct feature vectors $\mathbf{e}_{u\to v}$ and $\mathbf{e}_{v\to u}$, 
while the topology remains undirected. When branches rather than buses are the 
primary prediction target (as in line failure prediction or thermal limit monitoring), 
the line graph $\mathcal{L}(\mathcal{G})$ provides a natural alternative, 
converting each branch into a node and requiring a heterogeneous GNN to handle the 
resulting diversity of component types.

% For power systems, architecture choice has direct implications: long-range dependencies (e.g., cascading failures, wide-area stability) are poorly served by standard MPNNs, whose diffusive aggregation restricts information propagation across distant nodes~\cite{ringsquandlPowerRelationalInductive2021, pmlr-v251-nauck24a}; and edge tasks remain secondary in most standard architectures despite branches being the primary prediction target in many power system applications.

\paragraph{Limitations of MPNNs.}
Despite their success, MPNNs suffer from well-known failure modes. Oversmoothing refers to the tendency of node embeddings to become increasingly indistinguishable as the number of layers grows, as repeated neighborhood aggregation causes all nodes to converge to similar representations~\cite{arnaiz-rodriguezOversmoothingOversquashingHeterophily2026}. Oversquashing occurs when exponentially growing neighborhoods must be compressed into fixed-size embeddings, causing information from distant nodes to be lost~\cite{alonBottleneckGraphNeural2021}. Both phenomena limit the effective depth of MPNNs and motivate architectures with global receptive fields, such as graph transformers.

\subsection{Beyond MPNNs}
The limitations identified above, namely oversmoothing, oversquashing, and limited 
expressivity, have motivated a range of architectural extensions to the standard 
MPNN framework. This subsection also covers spectral methods, which predate 
message-passing but remain relevant as an alternative paradigm, before turning to 
extensions that address temporal dynamics, physical constraints, and domain-specific 
structure particular to power systems.

\paragraph{Spectral GNNs.}
GNN architectures are commonly divided into spectral and spatial categories. Spectral GNNs exploit the graph Laplacian eigenbasis to perform operations in the frequency domain, whereas spatial GNNs operate directly on local neighborhoods. In practice, the boundary is blurred: The graph convolutional network (GCN) by~\citet{kipfSemiSupervisedClassificationGraph2016}, though derived from spectral principles, reduces to simple 1-hop neighborhood aggregation in implementation, avoiding full Laplacian eigendecomposition. Consequently, works in the reviewed literature that describe their approach as spectral are in most cases employing this spatial approximation.

\paragraph{Graph Transformers.}
Graph Transformers (GTs) address MPNN limitations by attending to all nodes simultaneously rather than propagating information hop by hop, offering global receptive fields and greater representational capacity~\cite{mullerAttendingGraphTransformers2024}. The field is still emerging, with open questions in both theory and application. In power systems, architectures combining positional embeddings, GNN-based local aggregation, and global attention have been applied to tasks such as wind speed forecasting and false data injection attack detection~\cite{panShorttermWindSpeed2022, liDetectionFalseData2024}.

\paragraph{Physics-Informed GNNs.}
Incorporating domain knowledge such as physical laws and operational constraints 
into the model architecture improves performance and generalization while reducing 
the amount of training data required~\cite{misyrisPhysicsInformedNeuralNetworks2020}, 
and has been increasingly explored in recent years (Fig.~\ref{fig:gnn_trends}). 
For power systems, this means embedding constraints such as Kirchhoff's laws or 
power flow equations directly into the GNN, ensuring that predictions are 
physically consistent by construction rather than by coincidence.

\paragraph{Spatiotemporal GNNs.}
Many power system tasks involve signals that evolve over time across a fixed or slowly changing graph: Wind and solar forecasting, load prediction, and state monitoring are prominent examples. Spatiotemporal GNNs address this by coupling graph layers with temporal modules such as Recurrent Neural Networks (RNNs), Long Short-Term Memory networks (LSTMs), or temporal convolutional networks, effectively leveraging both structures simultaneously. This joint modeling allows the network to capture how conditions at one node evolve over time while remaining sensitive to correlated changes at neighboring nodes.

\paragraph{Power-grid-specific methods.}
\label{p:power-grid-specific}
Some works use power systems as a benchmark while focusing primarily on
methodological advances.
\citet{zhuScalingGraphNeural2024} investigate scaling behaviour and emergent
abilities of GNNs on large grids.
At the architecture level, \citet{plettenbergFlowAttentionalGraphNeural2025} enforce
Kirchhoff's current law within graph attention, while
\citet{pmlr-v251-nauck24a} introduce non-diffusive Dirac-Bianconi
GNNs that avoid oversmoothing over long ranges and operate on both node and edge features simultaneously, without reducing one to a function of the other. 
For dynamics,\citet{pmlr-v198-mukherjee22a} combine GNNs with Koopman operator theory to learn a distributed linearisation of sparse networked dynamical systems, exploiting network topology to improve both scalability and predictive performance.

% \paragraph{Non-black-box methods.}
% Both network science and simulation-based approaches provide non-black-box solutions that can be combined with GML methods. Network science applies graph-theoretic concepts to real-world networks, studying structural properties such as centrality, connectivity, and community structure. For example, in the work by~\citet{haqueGraphTheoryVs2024}, network science metrics are combined with time-domain discrete-event simulation to enable topology-informed security assessment. By leveraging measures such as betweenness centrality, eigenvector centrality, and edge betweenness, critical nodes and edges are identified and validated against simulation outcomes using real-world data. GML methods could also benefit from this information.

The architectures introduced above form the methodological foundation of this review. In practice, power system applications frequently also employ training paradigms such as reinforcement learning and federated learning that address control and privacy constraints respectively. Rather than treating these as isolated architectural variants, we discuss them in context within Section~\ref{s:applications}, where each arises naturally from the demands of a specific power system task.

\section{Graph Representation of Power Systems}
\label{s:power_grid_is_a_graph}

In the following, we detail how power system components and their physical interactions map onto graph representations, establishing the specific inductive biases that GML methods exploit.

\paragraph{Abstraction levels.} In standard formulations, buses, defined as points of interconnection between generators, loads, or substations, are represented as nodes, while branches (transmission lines and transformers) are represented as edges. However, this abstraction is itself a modeling choice, and the level of granularity can vary significantly depending on the task. At the finest level, every individual load, generator, and busbar is represented as a distinct node, preserving the full physical detail of the network. At an intermediate level, all devices connected to the same busbar are aggregated into a single node, which reduces graph size while retaining the essential connectivity structure. At the coarsest level, entire regions or subnetworks are collapsed into  single nodes, as is common in transmission planning studies where  computational tractability requires significant spatial aggregation. The choice of abstraction level fundamentally shapes which GML methods are appropriate, what features are available, and how well the resulting model generalizes. 

\begin{figure*}
\centering
\includegraphics[width=\linewidth]{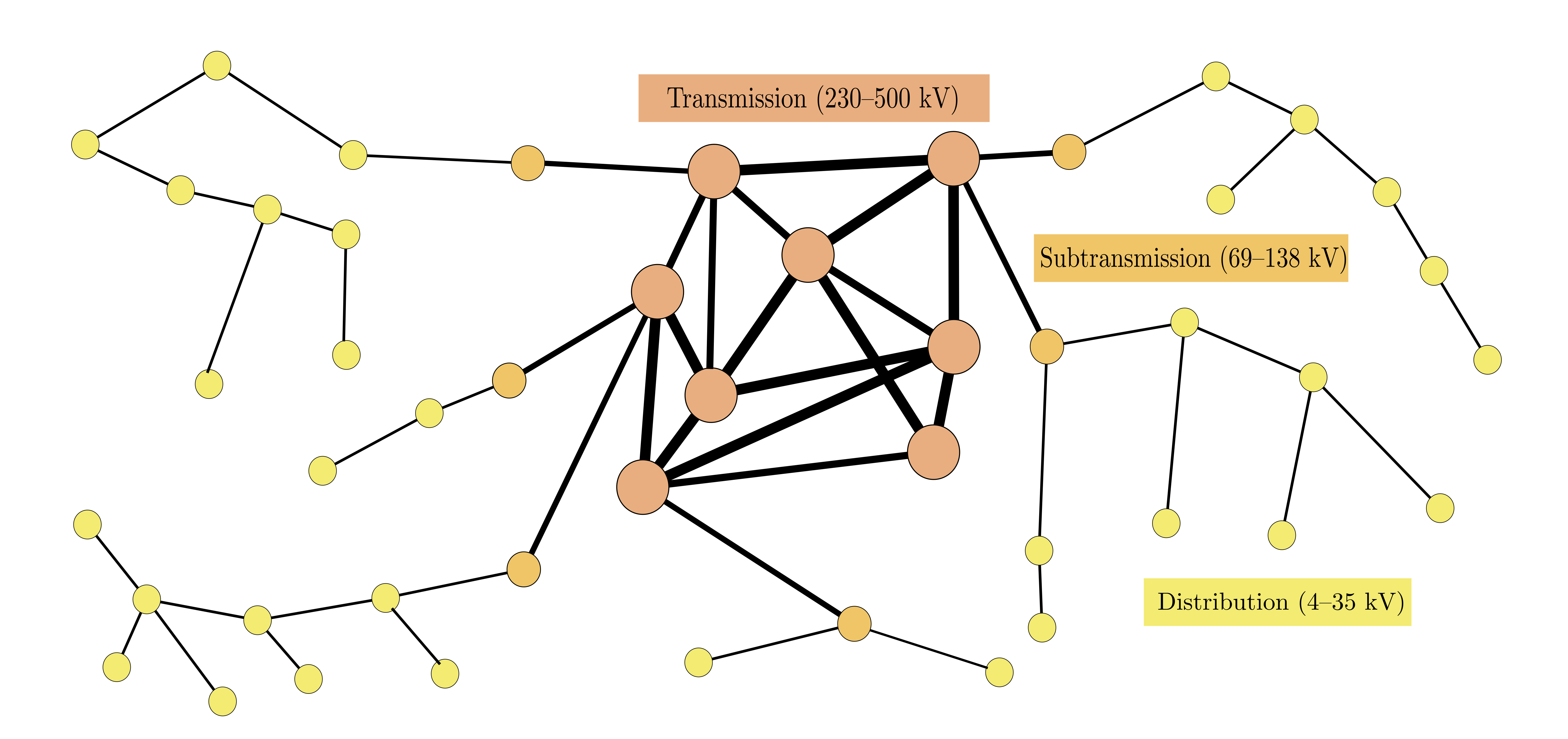}
\caption{Illustration of transmission, subtransmission, and distribution networks: differences in voltage levels, number of nodes, and graph topology (dense/meshed vs. radial/tree).}
\label{fig:transmission_distribution}
\end{figure*}

\paragraph{Transmission, subtransmission, and distribution.} Electric power systems vary significantly in size and structure. Typically, the network is divided into three main subsystems with decreasing voltage magnitude: transmission, subtransmission, and distribution (see Fig.~\ref{fig:transmission_distribution}). The transmission system transports power at the highest voltage across long distances and interconnects major generating stations with large load centers and transformers to lower voltage levels. The subtransmission system transfers power from transmission substations to distribution substations. In practice, the boundary between transmission and subtransmission is not always clearly delineated. Finally, the distribution system represents the last stage of electricity delivery for most commercial and residential customers, while large industrial consumers are typically connected directly at the transmission or subtransmission level. Overall, the system consists of multiple generating sources and several interconnected layers of transmission networks~\cite{kundurPowerSystemStability}. The topology of these subsystems differs fundamentally. Transmission grids are typically heavily meshed to ensure power can still be delivered after the outage of a single line, while distribution grids are operated mostly in a radial topology.

\paragraph{Scale and graph size.} The number of nodes depends heavily on the chosen level of abstraction. At the 
transmission level, European models such as PyPSA-Eur~\cite{horschPyPSAEurOpenOptimisation2018} 
operate at substation or bus level, yielding on the order of $10^3$--$10^4$ nodes, 
though planning studies commonly cluster these further to reduce computational cost. 
Distribution networks can in principle reach $10^7$--$10^8$ nodes when individual 
end users are included, but full topology at this resolution is rarely available, 
making this a fundamental data constraint rather than a computational one. In 
practice, studies focus on a single feeder or regional subnetwork, bringing the 
working graph to $10^2$--$10^4$ nodes. More broadly, the graph a practitioner 
works with is always a model of the grid, not the grid itself, and the choice of 
abstraction level fundamentally shapes which analytical methods are appropriate. 
The relevant question for GML is therefore the scale of the subgraph a given task 
requires, not the theoretical maximum scale of the grid. 

%Interconnection between neighboring systems improves both system security and economic efficiency ~\cite{kundurPowerSystemStability}. The European power system forms a single continental-scale synchronous grid, coordinated by the European Network of Transmission System Operators for Electricity (ENTSO-E). The Continental Europe Synchronous Area spans more than 20 countries and represents the largest interconnected grid worldwide, characterized by enormous operational complexity ~\cite{ucte2008}.

\paragraph{Beyond topology.} Power-grid topology matters, but topology alone is not sufficient.
%The power grid has also been investigated from the perspective of complex network science. ~\citet{paganiPowerGridComplex2013} systematically analyzed both real-world and synthetic topologies, reporting properties such as small-world structure and exponential degree distributions. More fundamentally, 
~\citet{hines2010topological} demonstrate that purely topological models are poor predictors of power grid behavior, as the physics of power flow cannot be captured by graph structure alone. \citet{paganiPowerGridComplex2013} reach the same conclusion in their survey, stressing that purely topological analysis is insufficient and that incorporating physical parameters such as line impedance is necessary for accurate vulnerability assessments. Recent work has begun connecting network-theoretic metrics, such as small-world structure and exponential degree distributions ~\citet{paganiPowerGridComplex2013}, with machine learning for power system tasks. For example, \citet{titz2024predicting} use interpretable gradient-boosted trees with static network features for stability prediction, while~\citet{zhu2025network} directly enrich GNNs with network measures.

% As outlined earlier, power system tasks can be formulated at three hierarchical levels: graph, node, and edge. Power system stability provides a representative example. At the graph level, the question is 
% global: is the system as a whole currently in a state prone to instability? At the node level, the question becomes local: would a fault originating at this generator or substation destabilize the 
% system? At the edge level, the focus shifts to individual branches: is this transmission line operating within its limits, or would its outage trigger a cascading failure? Each level corresponds to a different operational need, and the same underlying GML model can, in principle, produce outputs at all three simultaneously. This problem can be cast either as a regression task, where the objective is to predict a 
% continuous stability margin, or as a classification task, in which stability is categorized into binary or multiple discrete states. The subsequent sections examine how these task formulations appear across 
% the full range of power system applications.

\paragraph{Graph construction paradigms.}
Not all GML applications use the physical grid as their graph. In the first paradigm, 
the graph directly encodes physical topology: Nodes are buses, edges are lines, and 
the structure is given by the network itself. In the second, the graph is constructed 
to represent statistical or geographic relationships, as in forecasting applications 
where wind farms or PV plants are connected by spatial proximity or correlation. 
In such cases, graph construction is itself a modeling decision, and the resulting 
graph may bear no relation to the physical grid. This distinction recurs throughout 
the applications survey and has direct implications for what GML can and cannot 
contribute.

% 4. Applications
% Goal: Demonstrate value and state-of-the-art results.

% What is it about?
% Predictive: Load forecasting, renewable generation forecasting.
% Descriptive/State Estimation: Inferring grid state from sparse measurements.
% Prescriptive/Control: Optimal Power Flow (OPF), Fault location, Topology optimization.

% What is it not about?
% A long list of every paper ever written. Focus on archetypal problems that show GML's specific advantage over standard ML or traditional solvers.

% Audience Goals:
% PSE: Wants to see benchmarks against AC-OPF or Newton-Raphson. Did it actually work better/faster?
% ML: Wants to see novel applications of GNNs. Did they use attention mechanisms? Did they handle dynamic graphs?

\section{Applications}
\label{s:applications}

With the GML foundations and graph representation of power systems established, we now turn to applications of GML in power systems. Whether GML is suitable for a given power system application depends heavily on the problem context: power systems span an extreme range of timescales and operational regimes~\cite{machowskiPowerSystemDynamics2020}, from real-time control of fast dynamics to medium-timescale operational optimization and long-term planning, each posing methodologically distinct challenges that no single GML approach can address uniformly.

\begin{figure}
    \centering
    \includegraphics[width=\linewidth]{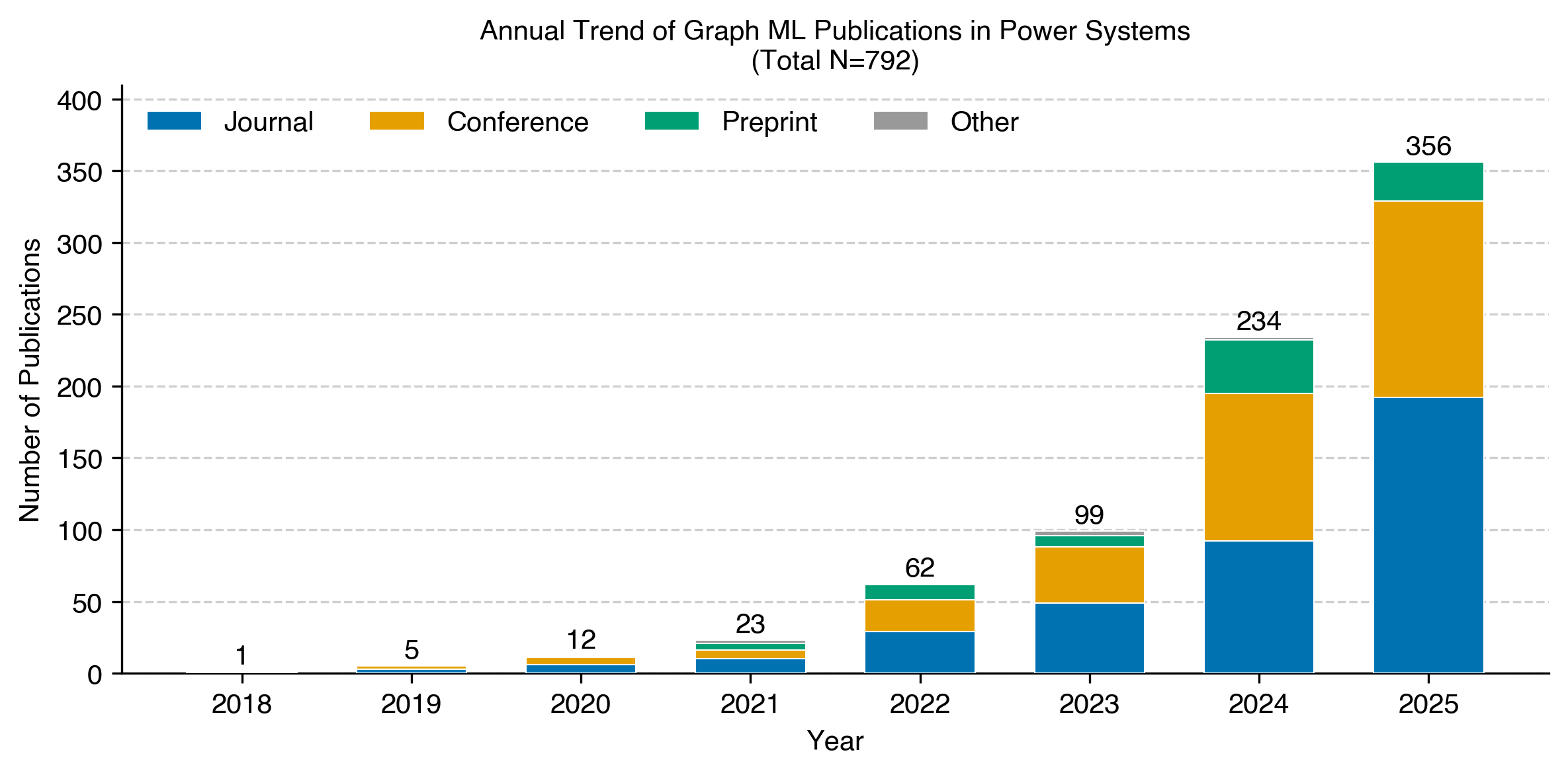}
    \caption{Annual GML publications in power systems (2018--2025), broken down by publication type.}
    \label{fig:publications_per_year}
\end{figure}

In the following, we present the results of our systematic survey along the main application areas in power systems. As shown in Figure~\ref{fig:publications_per_year}, the number of publications on GML for power systems has grown rapidly, reaching 234 in 2024 and 356 in 2025, bringing the total to 792 publications across all years, see also Methods for details on how we included papers. 

\begin{figure}
    \centering
    \includegraphics[width=\linewidth]{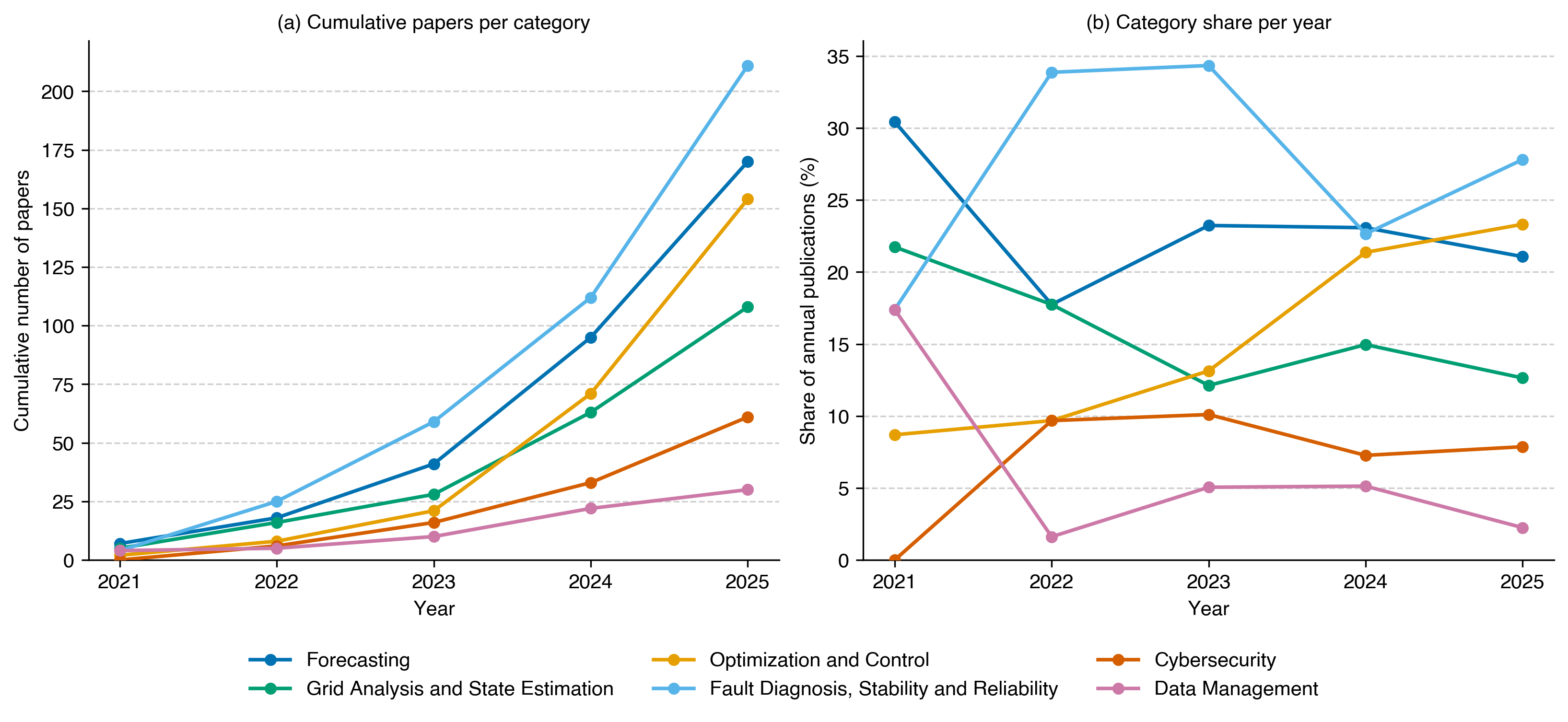}
    \caption{Cumulative publication counts per application category (left) and each category's annual share of total publications (right), revealing which areas have grown fastest since 2021.}
    \label{fig:topic_trend_category_trends}
\end{figure}

While the overall publication count for all topics has grown over time, we note that some topics have become more or less popular over time, see Figure~\ref{fig:topic_trend_category_trends}. E.g. the interest in "Data Management" dropped, while Forecasting and Fault diagnosis remained strong topics. An exhaustive discussion of all studies is neither  feasible nor instructive. For a comprehensive reference list, we refer the interested reader to our companion website. Meanwhile, here, we discuss what the general application challenges studied with GML and highlight a representative selection of papers, prioritizing methodological novelty and academic influence as reflected by citation metrics.

%A recurring observation across the literature is that GML models are reported to outperform baselines in the reviewed publications. This should be interpreted cautiously, however, as baselines frequently consist of non-graph ML models or simplified classical solvers rather than state-of-the-art domain methods, and the absence of shared datasets makes direct comparison across studies difficult.

\begin{figure*}
    \centering
    \includegraphics[width=0.8\linewidth]{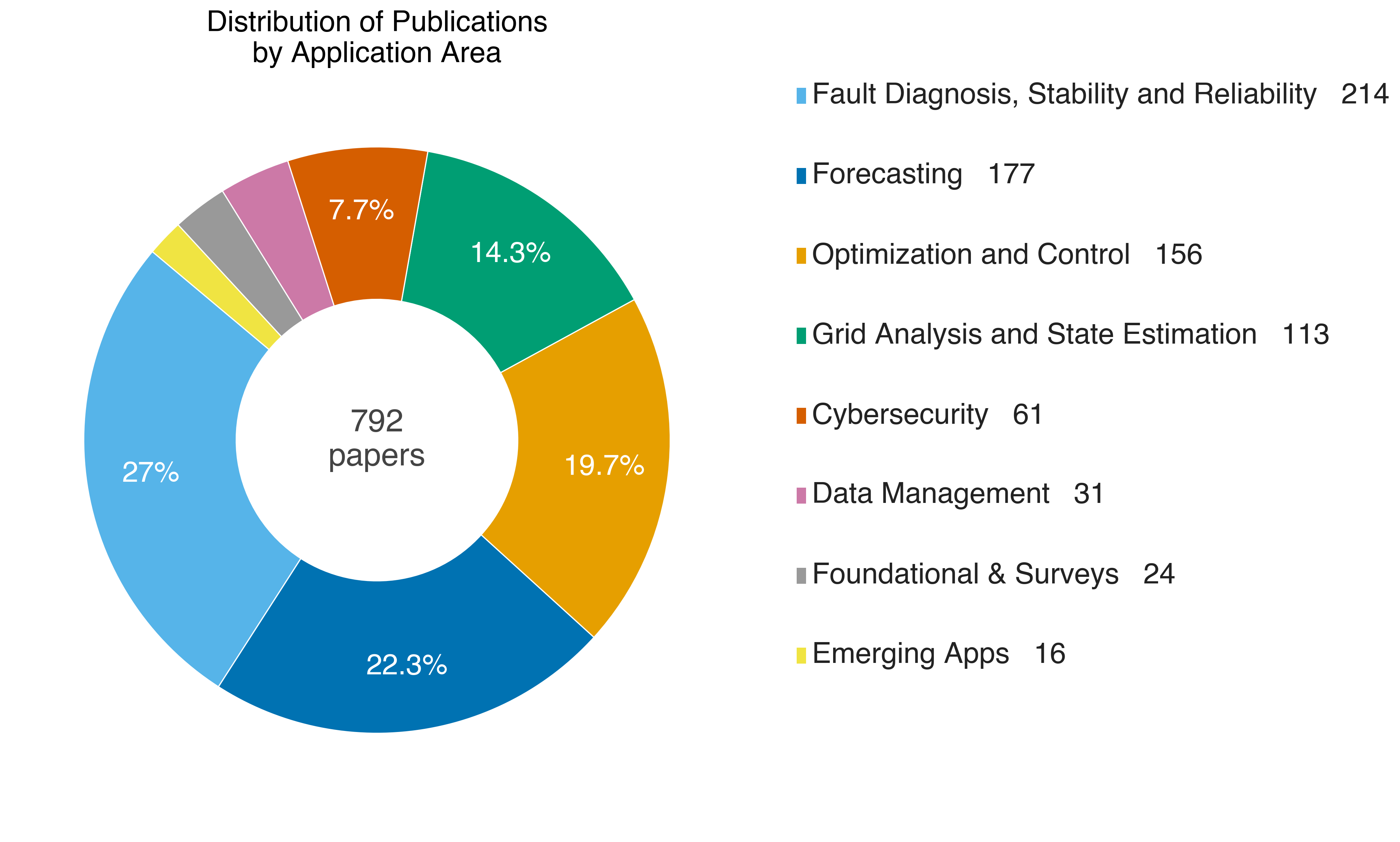}
    \caption{Distribution of 792 surveyed publications across GML application areas 
in power systems. The five largest categories are reviewed as dedicated subsections 
in Section~\ref{s:applications}. Data Management is discussed in 
Section~\ref{s:datasets_benchmarks}, Emerging Apps in 
Section~\ref{s:challenges_and_outlook}, and Foundational \& Surveys in 
Section~\ref{s:graph_machine_learning}.}
    \label{fig:applications_overview}
\end{figure*}

% thickness

We ordered the following subsections along the grid's operational logic: \textit{Observe} the current state, \textit{predict} future conditions, \textit{act} through optimization and control, \textit{defend} against faults and adversarial threats. Figure~\ref{fig:applications_overview} shows how many publications address each of these application areas.

% Application area 1
\subsection{Grid Analysis and State Estimation}
\label{ss:grid_analysis_state_estimation}
Grid analysis and state estimation is a group of inverse problems aiming to recover hidden system states or parameters from incomplete and noisy measurements. Specific examples include power system state estimation, power flow analysis, parameter identification, and topology identification, where physical laws act as hard constraints on any valid solution. GML are particularly well suited to respect the underlying physics, as it is already defined via the grid's topology. Thereby, a GML model can propagate information from observed to unobserved parts of the network in a way that is consistent with how the physical quantities themselves relate. 

\paragraph{State estimation.}
Power system state estimation aims to recover voltage magnitudes and phase angles from incomplete, noisy, and sometimes corrupted measurements. Classical weighted least-squares estimators and Kalman-filter-based variants rely on accurate network models, sufficient observability, and well-characterized measurement noise. These assumptions are increasingly strained in distribution grids, where sensing is sparse and measurements may be missing or unreliable. GML approaches successfully treat state estimation as a topology-aware inverse problem.
\citet{pagnierPhysicsInformedGraphicalNeural2021} propose Power-GNN, which embeds power flow constraints directly into the architecture and outperforms conventional neural networks on large-scale systems. A related line combines Kalman filtering with GNNs to capture spatiotemporal dependencies and estimate voltage magnitudes and phase angles using physics-consistent losses~\cite{ngoPhysicsinformedGraphicalNeural2024}. In distribution networks, \citet{habibDeepStatisticalSolver2024} show that GNNs constrained by power flow equations can learn directly from imperfect measurements, outperforming traditional methods that assume idealized input data. The main advantage of GML for state estimation is its ability to combine sparse measurements, topology, and physical constraints. The main open challenge is to provide robustness guarantees under topology errors, bad data, and distribution shifts.

\paragraph{Power flow analysis.} Traditional AC power flow solvers iteratively solve Kirchhoff's laws via Newton-Raphson methods to determine voltage amplitudes, angles, active and reactive power. These solvers are comparatively slow and become a bottleneck when thousands or millions of power flows must be evaluated, for example in N-2 contingency analysis or probabilistic security assessment,~\cite{dononNeuralNetworksPower2020}.
In this setting, GML acts as a fast surrogate for repeated solver calls rather than as an optimizer: the objective is to approximate physically consistent power flow solutions at much lower computational cost.
\citet{dononNeuralNetworksPower2020} propose a GNN-based method that accelerates computation by directly minimizing Kirchhoff's law violations at each bus during training, improving robustness to changes in injections, topology, and line parameters. This illustrates the main benefit of graph-based surrogates for power flow: they can exploit the locality of electrical coupling while remaining adaptable to changing grid structures. However, strict physical feasibility and reliable error bounds remain open issues.

\paragraph{Parameter and topology identification.}
Parameter and topology identification are closely related inverse problems: Parameter identification recovers unknown physical parameters of grid components, while topology identification infers the connectivity structure itself.
Classical approaches often rely on detailed measurements, probing signals, or repeated model fitting, which can be difficult in distribution systems with sparse observability.
GML is well positioned to address both problems because parameter identification can be cast as edge- or branch-level regression, while topology identification can be formulated as link prediction. AWL-GCN exploits neighboring branch information for branch parameter identification rather than treating each branch independently, and uses a self-tradeoff loss to remain resilient to noisy or divergent measurements~\cite{huangAWLGCNBranchParameter2023}. For topology identification, \citet{wangPowerSystemNetwork2021} combine a knowledge graph that mines candidate connections with a GNN that verifies them against measured features, improving robustness to missing or conflicting information. The main promise of GML in this area is its ability to reason jointly across the full graph parameters or candidate links. A key challenge is that the graph topology used within the GML itself is undetermined, making physical consistency hard to enforce.

%With a reliable picture of the current grid state established, we turn next to forecasting, where GML methods must predict future conditions without the graph being given by the physical network, making graph construction itself a modeling decision.

With a reliable picture of the current grid state established, we turn forecasting future grid states.% next to forecasting, where GML methods must predict future conditions without the graph being given by the physical network, making graph construction itself a modeling decision.

\subsection{Forecasting}
\label{ss:forecasting}
Forecasting is one of the most active areas of GML applications in power systems by publication volume(see Figure~\ref{fig:topic_trend_category_trends}), yet it differs fundamentally from grid analysis: In many forecasting tasks, the graph is not given directly by the electrical network. Instead, it must be constructed from geographic proximity, weather similarity, statistical correlation, or learned dependencies between sites. This makes graph construction a central modeling decision. %: a poorly chosen graph can encode misleading correlations, while a well-chosen graph allows the model to exploit spatial dependencies that independent time-series models cannot capture.
The main forecasting subdomains are wind power and wind speed forecasting~\cite{yuSuperpositionGraphNeural2020, khodayarSpatioTemporalGraphDeep2019}, photovoltaic (PV) power forecasting~\cite{simeunovic_spatio-temporal_2022}, and residential or regional load forecasting~\cite{linSpatialTemporalResidentialShortTerm2021}. %Across these tasks, GML typically acts as a spatiotemporal dependency learner: graph layers encode spatial relationships across sites, while recurrent, convolutional, attention-based, or transformer components model temporal dynamics.

\paragraph{Wind and PV forecasting.}
Wind and PV forecasting aim to predict renewable generation over short- to medium-term horizons. Classical statistical models and site-independent neural networks can capture local temporal patterns, but they often fail to exploit correlations between neighboring plants or sites exposed to the same weather systems. GML addresses this by representing wind farms or PV plants as nodes and connecting them through geographic proximity, meteorological similarity, or learned spatiotemporal correlations.
For wind forecasting, graph-based models have been used to capture regional dependencies between wind farms and improve robustness to local anomalies~\cite{khodayarSpatioTemporalGraphDeep2019, wangDynamicSpatiotemporalCorrelation2022, panShorttermWindSpeed2022}. PV forecasting follows a similar logic: graphs are constructed from geographic and weather-related features to provide site-specific context that global aggregation loses, with reported advantages especially at longer forecast horizons~\cite{karimiSpatiotemporalGraphNeural2021, zhangOptimalGraphStructure2023}. The main advantage of GML in renewable forecasting is its ability to exploit spatial weather-driven dependencies, while a key challenge is to construct a useful graph from the available data.

\paragraph{Load forecasting.}
Load forecasting predicts future demand at the household, feeder, bus, or regional level. Classical approaches include for example autoregressive models, recurrent or convolutional neural networks. 
GML offers a consistent way to include spatial dependencies in different load time series, e.g. due to temperature or humidity driving demand or similar dynamics caused by socioeconomic patterns or distribution grid topology.
\citet{linSpatialTemporalResidentialShortTerm2021} show how graph-based models capture spatial correlations in residential demand driven by common exogenous factors. With the increasing availability of smart-meter data, GML methods can also incorporate richer node-level features for bus- or customer-level forecasting~\cite{huangGatedSpatialtemporalGraph2023}. Multi-energy forecasting extends this setting by coupling electricity, heat, gas, and other energy carriers, introducing irregular time steps and multiscale spatial correlations; spatiotemporal GNNs have shown advantages over non-graph architectures in such settings~\cite{zhuangMultiScaleSpatialTemporalGraph2024}. A key open challenge is to construct graphs that generalize across seasons, customer populations, and changing consumption behavior.

%Accurate forecasts of future conditions set the stage for the next challenge: using that information to operate the grid as efficiently and safely as possible through optimization and control.

Accurate forecasts set the stage to optimizing and controlling grids.

% Application area 3
\subsection{Optimization and Control}
\label{ss:optimization_and_control}
The third application area comprises power system optimization and control. Representative problems include AC/DC optimal power flow (OPF), unit commitment, topology optimization, and voltage control. The key distinction from forecasting and state estimation is that outputs must be not only accurate but also feasible: solutions must satisfy hard physical, operational, and security constraints. %An infeasible control action is not merely imprecise but operationally unusable. 
GML therefore acts as a solver surrogate, solver accelerator, or policy representation, rather than merely as a predictor.

\paragraph{Optimal power flow.} Optimal power flow (OPF) describes a group of constrained optimization problems that are essential for the operation of modern power grids. Classical AC optimal power flow (AC-OPF) seeks to minimize generator costs while satisfying operational and security constraints, resulting in a non-linear and non-convex optimization problem that is strongly NP-hard~\cite{bienstock2019strong}. Hence, practical applications often resort to linear power flow approximations, but this can seriously impede the accuracy of the solutions~\cite{coffrin2014primal}. Machine learning-assisted OPF (MLOPF) aims to accelerate solution times by shifting the computational burden from online optimization to offline model training, followed by fast online inference. Due to the importance of OPF for power grid operation, interest in MLOPF methods, including both graph-based and non-graph-based approaches, has been growing~\cite{khaloieReviewMachineLearning2025}. These approaches can be categorized into two groups: End-to-end (E2E) models learn a direct mapping from input to output, while Learning-to-Optimize (L2O) approaches leverage machine learning to assist traditional solvers.

A simple approach for E2E MLOPF is imitation learning, where a direct input-to-solution mapping is learned to replicate the outputs of a traditional solver~\cite{owerkoOptimalPowerFlow2020}. However, by neglecting feasibility constraints, it may produce solutions that are not AC-feasible or N-1 secure. The N-1 criterion requires the system to remain within operational limits after the outage of any single relevant component.

One option to improve the feasibility of E2E OPF is to include an appropriate regularization~\cite{liuTopologyAwareGraphNeural2023}. To obtain feasible solutions, it has been suggested to use GNNs to generate a warm start for a conventional optimizer, which then refines the solution to full feasibility~\cite{baker2019learning}. L2O approaches include models that identify non-binding constraints and thereby allow the traditional solver to speed up by solving a reduced OPF problem~\cite{phamN1ReducedOptimal2024}.

\citet{falconerLeveragingPowerGrid2023} compared GNNs to CNNs and Fully-Connected Neural Networks in E2E as well as L2O approaches. Their work showed that for fixed-topology OPF problems, simple fully connected networks are sufficient. Still, GNNs outperform other models in variable-topology scenarios due to their ability to adapt to structural changes.

Recently,~\citet{pilotoCANOSFastScalable2024} proposed a GNN-based method that provides near-optimal and near-feasible solutions and is robust to N-1 perturbations. While previous work mostly used smaller IEEE test cases, they showed a sub-linear scaling with grid sizes up to 10,000 buses. The authors of this study published their dataset OPFData~\cite{lovettOPFDataLargescaleDatasets2024} of solved AC-OPF problems with N-1 perturbations, allowing for transparent comparison across ML architectures.

The key advantage of graph-based models for OPF is their ability to naturally model the grid topology and thereby be robust to contingencies or generalize to different grids. Open challenges remain in ensuring strict feasibility while retaining optimality and investigating methods for multi-period OPF, topology switching (optimizing line on/off states as binary decision variables), and the Unit Commitment Problem. 

\paragraph{Reinforcement learning.}
GNNs combined with reinforcement learning have emerged as a promising approach for power grid control, where GNNs encode the graph-structured grid state to enable RL agents to learn control policies that generalize across unseen topologies, an advantage consistently demonstrated over non-graph-based alternatives across applications including topology optimization and voltage regulation~\cite{hassounaGraphReinforcementLearning2026}. As shown in Figure~\ref{fig:gnn_trends}, reinforcement learning appears in approximately 5\% of GML-based power system publications in recent years, reflecting its specialization to sequential decision-making tasks such as real-time grid control. However, these approaches remain largely at proof-of-concept stage, limited by small benchmark grids and the absence of standardized evaluation protocols~\cite{hassounaGraphReinforcementLearning2026, marchesini2025rl2grid}

\paragraph{Unit commitment.}
Unit commitment determines which generators should be turned on or off over a future time horizon while satisfying e.g. demand and network constraints~\cite{abujaradRecentApproachesUnit2017}. 
%Compared with OPF, unit commitment introduces discrete decisions and temporal coupling, making the problem computationally challenging, especially under high renewable penetration and increasing uncertainty. 
GML can support unit commitment either by predicting high-quality commitment schedules directly, screening candidate decisions, or providing confidence estimates that guide downstream optimization.
\citet{parkConfidenceAwareGraphNeural2024} use GNNs to predict generator commitments and active transmission lines with associated confidence scores, producing high-quality feasible schedules while highlighting the importance of uncertainty quantification in ML-based grid operations. %This illustrates a broader role for GML in mixed-integer grid optimization: graph models can exploit network structure to reduce the effective search space, but they do not yet replace the feasibility and optimality guarantees of established mixed-integer optimization methods.

\paragraph{Voltage control.}
Voltage control aims to keep voltage magnitudes within operational limits. The problem is particularly important in distribution systems with high PV penetration, where local voltage violations can arise rapidly and observability is often limited. Classical rule-based or optimization-based controllers may struggle to coordinate many distributed devices under changing topology and uncertain measurements. GML addresses this by encoding the electrical neighborhood of each controllable device, while reinforcement learning or other control methods learn actions on top of these topology-aware representations.
\citet{wangGraphLearningBasedVoltage2024} combine GNNs with reinforcement learning for cost-effective voltage control across multiple microgrids, enabling decentralized and privacy-preserving training. 
Physics-informed GNN-based approaches have further been proposed to stabilize training and improve sample efficiency in reactive power control by PV inverters~\cite{zhangPhysicsInformedMultiAgentDeep2024}.
The main advantage of GML for voltage control is its ability to coordinate spatially distributed actions, while particularly dealing with unseen operating conditions and ongoing topology changes are open challenges.

% \paragraph{Other.} GML has also been applied to controlled islanding, where the grid is deliberately partitioned into isolated subsystems to arrest cascading failures~\cite{dingControlledIslandingPower2024}, a problem that combines elements of topology optimization and real-time control.

A shared open problem across these tasks is providing formal feasibility and optimality guarantees, which classical solvers offer by construction but GML methods currently do not. %Even with optimal control strategies in place, faults and disturbances are inevitable. The ability to rapidly detect, localize, and respond to failures is therefore essential for maintaining grid reliability, which we discuss in the following section.

% Application area 4

\subsection{Fault Diagnosis, Stability and Reliability}
\label{ss:fault_diagnosis}
As part of critical infrastructure, power systems must remain reliable under a wide range of operating conditions and contingencies. This section groups three related but distinct task types. \textit{Fault diagnosis and localization} detects and pinpoints equipment failures or line faults, often under real-time constraints. \textit{Transient stability assessment} determines whether the system will remain synchronous after a large disturbance, a task traditionally addressed through computationally expensive time-domain simulation. \textit{Reliability and risk assessment} operates on longer timescales, estimating the likelihood of failures, cascading outages, or security violations under uncertain operating conditions.

\paragraph{Fault diagnosis and localization.}
Fault diagnosis and localization aim to detect abnormal events and identify the failed component as quickly as possible. Classical approaches rely on signal processing, or model-based residuals, but their performance can degrade under noisy measurements, missing data or changing topology. GML frames fault localization as a node- or edge-level classification problem: Because disturbances propagate through physical connections, the graph structure directly encodes the spatial pathways along which fault signatures spread.
\citet{chenFaultLocationPower2020} demonstrate this on the IEEE 123-bus system, showing that a GCN retains high accuracy under measurement noise and data loss and generalizes to unseen topology changes without retraining. A key limitation is the scarcity of labeled real fault data, especially for distribution-system events.

\paragraph{Transient stability assessment.}
Transient stability assessment determines whether generators remain synchronized following a large disturbance such as a short circuit, line outage, or sudden loss of generation. Classical assessment relies on time-domain simulation, which can be computationally expensive. GML provides a fast surrogate for dynamic security assessment by learning spatiotemporal representations of post-disturbance trajectories.
Transient stability assessment can be cast as graph-level classification of post-disturbance stability, regression of a stability margin, or node-level identification of critical generators and vulnerable areas. \citet{zhuIntegratedDataDrivenPower2024} use a GNN-based framework to extract wide-area spatiotemporal features for real-time stability monitoring and emergency control. This supports targeted corrective actions, such as identifying critical generators, and demonstrates greater computational efficiency and adaptability compared to classical methods. Again, a key challenge is the scarcity of realistic benchmark datasets, ideally based on real events.

\paragraph{Reliability and risk assessment.}
Reliability and risk assessment operate on longer timescales than fault diagnosis, estimating whether the grid can withstand contingencies or cascading events under uncertain operating conditions. Classical approaches rely on contingency screening, mathematical optimization, or repeated simulation. These methods provide strong physical grounding but can become computationally expensive when many operating points, outages, or cascading scenarios must be evaluated. GML addresses this by learning topology-aware classifiers or regressors that approximate the outcome of contingency or cascading-failure analyses.
For N-1 contingency assessment, \citet{cambiervannootenGraphNeuralNetworks2025} propose a GIN-based framework that classifies whether a medium-voltage grid satisfies the contingency criterion, achieving up to 1000 times faster assessments than mathematical optimization approaches while generalizing to unseen grid topologies. At a finer granularity, \citet{bhailaCascadingFailurePrediction2024} predict the post-event failure status of individual buses and branches following a cascading event, framing the problem as simultaneous node and edge classification and showing that incorporating branch features directly into message passing improves prediction over node-only approaches. A central challenge is to combine fast graph-based risk screening with 100\% reliable models.

%Beyond equipment failures and natural disturbances, modern grids face a distinct threat class: deliberate adversarial attacks on measurement and control infrastructure. We turn next to how GML addresses these cybersecurity challenges.

Havoing covered failures, we next turn to deliberate adversarial attacks and how GML addresses such cybersecurity challenges.

%The key advantage of GML for fault diagnosis is rapid topology-aware localization; the main limitation is the lack of realistic labeled fault data and validated dynamic benchmarks.

\subsection{Cybersecurity}
\label{ss:cybersecurity}
Unlike the physical faults and disturbances discussed in the preceding section, cybersecurity threats are human-made and adversarial. Attackers deliberately manipulate measurements, communication channels, or control signals in ways designed to mimic normal operating behavior, requiring detection methods that go beyond conventional physical anomaly recognition~\cite{khurana2010smart, yan2012survey}. %For GML, the central opportunity is that cyberattacks on power systems are constrained by both the communication infrastructure and the physical grid: malicious measurements must remain plausible with respect to neighboring measurements, power flow relations, and topology.

%\paragraph{False data injection attacks.}
False data injection attacks (FDIAs) compromise measurement integrity by injecting malicious data into smart meters, phasor measurement units, or state-estimation inputs while attempting to bypass classical bad-data detection systems~\cite{khurana2010smart}. Traditional residual-based detectors can fail when attacks are constructed to be consistent with the assumed measurement model. GML-based detectors utilize message passing to reveal inconsistencies between local and propagated measurements.
\citet{boyaciGraphNeuralNetworks2022} propose a GNN-based detector that exploits spatial correlations of measurements across the grid topology, outperforming existing detectors on IEEE 14-, 118-, and 300-bus systems while scaling linearly with network size. 

A related challenge is attack localization under incomplete topological information. \citet{quLocalizationDummyData2024} address this with a spatiotemporal graph wavelet approach that adaptively captures correlations between measurement data and grid topology, enabling rapid and accurate localization of attack origins even when full topology is unavailable. \citet{haghshenasTemporalGraphNeural2023} further demonstrate that temporal GNNs can detect both FDIAs and ramp attacks with high accuracy on simulation data. An open challenge is adversarial robustness: Because attackers can also use ML and adapt to learned detectors, future GML cybersecurity methods must be evaluated against adaptive attacks.

% Application area 6 (Data Management moved to Datasets and Benhmarks)

Across all application areas surveyed, progress is fundamentally shaped by the availability and quality of data. The following section examines the current state of datasets and benchmarks for GML in power systems, the structural constraints that limit data availability, and the methods the community has developed to work within them.

\section{Datasets and Benchmarks}
\label{s:datasets_benchmarks}
Credible progress in GML for power systems depends on shared datasets and transparent benchmarks. Without them, results are hard to verify, compare, and build upon. Most papers apply GML to a concrete domain problem under specific constraints, often comparing against non-graph baselines rather than on standardized datasets. Only a small fraction of publications provide open access to code, datasets, or trained models. Many mention only dataset names or minimal preprocessing/hyperparameter details. This lack of transparency impedes verification, slows innovation, and reduces trust in findings. We urge for more openly shared data and provide guidelines for GML-ready benchmark datasets at the end of this section.

\paragraph{Data scarcity.}
The data challenges in power systems are severe and deserve explicit attention. 
High-fidelity models of real systems are rarely accessible for methods research, 
and generating large datasets through simulation is computationally prohibitive. 
Crucially, no simulation framework comes close to jointly capturing all relevant 
factors (consumer behavior, machine dynamics, weather drivers, fault and protection 
mechanisms) at high fidelity. Empirical, operational data exists only 
in limited quantities at the transmission level and is effectively absent at lower levels (e.g. distribution), 
partly due to privacy and security constraints that preclude public disclosure. 
In practice, sensing gaps are additionally common due to failures, data injection attacks, and unreliable communication~\cite{wuMissingDataRecovery2021}, meaning that even available data is often incomplete or unreliable. At the distribution 
level, even basic grid topology is often not reliably known, and operational 
decisions are typically made by expert judgment rather than model-based 
optimization, leaving standardized evaluation protocols largely absent. This 
structural data scarcity is not a temporary limitation to be resolved by more 
data collection, it is a fundamental constraint that GML methods must be designed 
to handle.

\paragraph{GML addressing data scarcity.}
Several GML methods directly target the data constraints described above. 
Missing data imputation exploits network-mediated correlations between sites 
rather than treating each time series independently, significantly improving 
recovery of missing smart meter 
readings~\cite{kuppannagariSpatioTemporalMissingData2021}. Beyond processing 
available data, GML also offers tools for generating realistic synthetic data: 
GAN-GCN hybrids have been used to produce synthetic feeders indistinguishable 
from real node-level data~\cite{liangFeederGANSyntheticFeeder2021}, and graph 
autoencoders can learn generative distributions over spatiotemporal grid 
data~\cite{khodayarConvolutionalGraphAutoencoder2020}, pointing toward a future 
where GML reduces dependence on expensive physics-based simulation for dataset 
generation. Transfer learning bridges the simulation-to-reality gap by 
pre-training on abundant simulated data and fine-tuning with limited real 
measurements, with domain adaptation techniques mitigating distribution 
shift~\cite{pourmoradiMultitaskActiveTransfer2025, wuEfficientResidentialElectric2023}. 
Finally, federated learning addresses privacy constraints by allowing GNNs to 
learn cross-utility patterns while keeping sensitive operational data on 
premises~\cite{liFederatedGraphLearning2024}, enabling training across 
geographically distributed grids without centralizing topology, load, or 
event logs.

\begin{table*}[ht]
\centering
\caption{Publicly available benchmark datasets for GML in power systems. Abbreviations: NREL = National Renewable Energy Laboratory; TAMU = Texas A\&M University; EPA = U.S. Environmental Protection Agency; EIA = U.S. Energy Information Administration; L2RPN = Learning to Run a Power Network.}
\begin{tabular}{lp{10cm}}
\toprule
\textbf{Dataset} & \textbf{Applicable Tasks} \\
\midrule
\href{https://matpower.org}{\textbf{IEEE Standard Bus Systems}} & Power flow analysis, Optimal Power Flow (OPF), fault location estimation, cascading failure simulation, state estimation, transient prediction, voltage fluctuation prediction \\
\midrule
\href{https://figshare.com/articles/dataset/PowerGraph/22820534}{\textbf{PowerGraph}}~\cite{varbellaPowerGraphPowerGrid2024a} & Power flow, OPF, cascading failure prediction, voltage stability monitoring \\
\midrule
\href{https://figshare.com/projects/SafePowerGraph_-_NDDS25/212777}{\textbf{SafePowerGraph}}~\cite{ghamiziSafePowerGraphSafetyawareEvaluation2024} & Safety-critical power system operations, topology-aware forecasting, N-1 secure OPF, fault localization \\
\midrule
\href{https://simbench.de}{\textbf{Simbench}}~\cite{meineckeSimBenchBenchmarkDataset2020} & Comparative benchmarking of power flow and OPF solutions in realistic distribution and transmission grids \\
\midrule
\href{https://www.nrel.gov/grid/solar-power-data.html}{\textbf{NREL PV Datasets}}~\cite{draxlOverviewMeteorologicalValidation2015} & Multi-site photovoltaic (PV) power forecasting, handling missing data, renewable integration studies \\
\midrule
\href{https://electricgrids.engr.tamu.edu/}{\textbf{ACTIVSg / TAMU Test Cases}} & Large-scale power flow, OPF, and topology-aware tasks; synthetic grids (200--70{,}000 buses) statistically similar to real U.S. networks, with geographic coordinates and realistic load/generation profiles \\
\midrule
\href{https://egriddata.org/}{\textbf{eGridData}} & Load forecasting, renewable integration, and emission-aware dispatch studies using aggregated real-world grid data (EPA eGRID, EIA) \\
\midrule
\href{https://lfenergy.org/projects/grid2op/}{\textbf{Grid2Op / RL2Grid}}~\cite{marchesini2025rl2grid} & Topology optimization and sequential decision-making via reinforcement learning; basis of the L2RPN competition series; GNN-based policies identified as a future direction \\
\bottomrule
\end{tabular}
\label{tab:gnn_power_datasets}
\end{table*}

\paragraph{Public benchmark landscape: strengths and limits.}
While GML-based approaches to data generation and management show promise, 
they remain largely at the proof-of-concept stage and cannot yet substitute 
for well-curated reference datasets. Credible progress in GML for power systems 
therefore continues to depend on shared datasets and transparent benchmarks.
IEEE test cases (e.g., 14-, 39-, 118-bus) are widely used standardized systems originally designed to validate power-flow algorithms. They enable controlled, reproducible experiments for optimization, control, and learning. However, they are limited in scale and do not reflect the diversity and dynamics of modern grids (renewables, demand response, cyber-physical constraints). Larger publicly documented systems exist, e.g., the Polish system~\cite{Polish27362736} (as of 2004) with 2{,}736 nodes and the PEGASE European system~\cite{Pegase9241} with 9{,}241 nodes.
%and operational grids can exceed 100{,}000 buses~\cite{safdarianPowerSystemSparse2022}.
As GML moves beyond early demonstrations, more realistic and larger datasets are needed. The ACTIVSg synthetic test cases (see Table~\ref{tab:gnn_power_datasets}), ranging from 200 to 70{,}000 buses and designed to be statistically similar to real U.S. networks with geographic coordinates, represent a step in this direction. Beyond scale limitations, these benchmarks exclusively reflect the infrastructure  of industrialized nations. Power systems in developing regions differ fundamentally  in topology and operational regime, and remain entirely absent from the GML literature.

% \paragraph{Synthetic vs.\ real data and access constraints.}
% Studies often rely on synthetic data generated under simplified or unspecified assumptions. While useful for controlled experiments, such data rarely captures real-world variability. In practice, sensing gaps are common due to failures, data injection attacks, and unreliable communication~\cite{wuMissingDataRecovery2021}. Even though power systems generate large volumes of real-time data, access is often restricted by privacy, security, and resource constraints~\cite{ngoPhysicsinformedGraphicalNeural2024}. Public datasets typically lack system-wide coverage or sufficient resolution~\cite{varbellaPowerGraphPowerGrid2024a}. When real data are scarce, realistic synthetic datasets and transfer learning can help; for instance, ~\citet{linResidentialElectricLoad2021} improve residential load forecasting with an attentive transfer framework for GML.

\paragraph{Existing efforts and gaps.}
~\citet{varbellaPowerGraphPowerGrid2024a} argue that public power-grid datasets such as the Electricity Grid Simulated, PSML (Power Systems Machine Learning), and Simbench datasets are not tailored for graph machine learning. They propose what is, to our knowledge, the most suitable benchmark to date for cascading-failure and power-flow tasks, including explainability aspects. Beyond static datasets, the Grid2Op framework~\cite{marchesini2025rl2grid}, developed under LF Energy and underlying the L2RPN competition series, provides a simulation environment for sequential topology optimization and GNN-based control policies have been identified as a promising future direction. Additional datasets and environments are listed in Table~\ref{tab:gnn_power_datasets}.

The datasets listed above reveal an uneven coverage across task types.
For static tasks such as power flow and optimal power flow, the situation 
is relatively mature: Multiple benchmark systems exist at various scales, 
and simulation-based dataset generation is straightforward since a single 
steady-state solve is computationally cheap. The picture is far more 
challenging for dynamic tasks. Fully dynamic grid models that capture 
electromechanical transients, protection system responses, and 
inverter-based resource behavior are rare, and generating large-scale 
labeled datasets from time-domain simulation is computationally 
prohibitive. To our knowledge, no publicly available dataset provides 
the combination of realistic dynamic grid models, diverse fault scenarios, 
and sufficient scale required to train and benchmark GML models for 
transient stability or fault analysis in a rigorous way. This gap is 
particularly concerning given that dynamic security assessment is among 
the most safety-critical applications motivating GML adoption in power 
systems.

\paragraph{Toward GML-ready benchmarks.}
To address these shortcomings, we argue that benchmarks should satisfy 
the following criteria:
\begin{enumerate}
    \item \textbf{Explicit graph representation.} Datasets should be released in a standardized graph format with clearly defined node and edge semantics, including physical interpretations of features such as bus type, line impedance, and injection profiles.
    \item \textbf{Standardized evaluation protocol.} Benchmarks should specify canonical train, validation, and test splits and preprocessing pipelines. Accompanying papers should additionally report hyperparameter search ranges and selection criteria, so that results can be reproduced without access to unpublished implementation details.
    \item \textbf{Task and metric specification.} Each benchmark should define concrete tasks (e.g., AC power flow, N-1 security assessment, cascading failure prediction) alongside domain-appropriate metrics (e.g., power flow residual, voltage violation rate, stability margin).
    \item \textbf{Graph and non-graph baselines.} Classical solvers (e.g., Newton--Raphson, interior-point OPF) and non-graph ML baselines should accompany GNN baselines, enabling fair assessment of the marginal benefit of graph inductive biases.
    \item \textbf{Reproducibility infrastructure.} Data loaders, graph construction, training, and evaluation code should be released within common frameworks, together with trained model checkpoints and fully specified computational environments.
    \item \textbf{Physical and operational realism.} Datasets should reflect realistic conditions, including load variability, renewable intermittency, topology changes, and sensing imperfections, which are underrepresented in current benchmarks.
    \item \textbf{Licensing and documentation.} Clear licenses, provenance statements, and usage documentation are essential for legal reuse and broader community adoption.
\end{enumerate}

%\textbf{Recommendations.}
The recent work by ~\citet{feyPyG20Scalable2025} highlights three key aspects of the recent PyG framework extensions: support for heterogeneous and temporal graphs, scalability, and explainability. These aspects are directly aligned with power system applications, which are inherently large-scale, multi-modal, dynamic, and safety-critical. We advocate mandatory sharing of code and trained models alongside publications, and recommend building on established libraries (e.g., PyG) to increase reproducibility and lower engineering overhead.

% 5. Outlook and Challenges
% Goal: Set the research agenda for the next 5 years.

% General Problems:
% Scalability: Can GNNs handle 100,000+ bus systems?
% Explainability: Can a grid operator trust a "black box" GNN during a blackout?
% Oversmoothing (and other common GML problems): A common GML problem where deep networks make all node representations identical. Does this happen in large grids?

% Audience Goals:
% Shared: Both groups need to solve the "Trust" problem. How do we verify ML outputs satisfy safety constraints?

% END EVERY SECTION (PARAGRAPH) WITH OUTLOOk/CALL TO AN ACTION/NEXT?

\section{Challenges and Future Directions}
\label{s:challenges_and_outlook}

The application survey in Section~\ref{s:applications} and the benchmark assessment in Section~\ref{s:datasets_benchmarks} together reveal a consistent pattern: GML methods frequently demonstrate strong performance in controlled settings, yet exhibit limited transferability to operational environments. In this section, we first discuss which application areas for GNNs are emerging and should be investigated in the future, then examine challenges intrinsic to GML that manifest in power system contexts, and finally turn to the domain-specific barriers that distinguish power grids from other graph learning applications.

% Everything else, that couldn't fit any of other categories/emerging
\paragraph{Emerging and future applications.}
This review focuses on work published until the end of 2025. For more recent developments, we refer to our companion webpage \url{https://kit-iai-dracos.github.io/graph-ml-power-sys/}, which we update regularly. We note that we intentionally focused on the technical aspects by searching for power systems, see also Methods, while economic aspects of energy systems, such as trading across multiple zones, may not be fully covered as authors might use different terminology.

That said, existing emerging applications include electricity markets~\cite{huangGraphConvolutionalSpectral2024, shanMultidimensionalComprehensiveEvaluation2024}, transmission and grid planning \cite{liAdaptiveKnowledgeAssessment2024, liuConstructionKnowledgeMap2025, mujicaHybridRNNGNNApproach2025}, carbon emission monitoring \cite{zhuTwoStageRealTimeCarbon2025}, electric vehicle integration \cite{natrayanDCMicrogridsAPDGNNALO2025}, and operator training systems \cite{heManualOperationEvaluation2022}.
While the current list of GML applications in power systems is already impressive, we identify several areas on which additional focus could be spent. 

First, most GNN studies focus on individual power system aspects, such as generation or load in isolation. Meanwhile, future power systems will display an increasing number of prosumers who can shift their demand and act as consumers or generators at different times of the day. Furthermore, sectors will be coupled: Heating, transportation, and electricity will need to be modeled together. We suspect that such integrated approaches will be challenging but substantially improve applicability.

Second, and closely related, current power system practice mostly relies on simulations and conventional optimization, while GML approaches attempt to solve these challenges purely via data. We suggest developing hybrid approaches that combine the reliability of physics-based simulations with the representational capabilities of GML. This also implies addressing the static graph assumption common in current models: Real-world grids evolve over time through expansion, faults, and reconfiguration, and hybrid frameworks are a natural place to incorporate continual or topology-aware learning.

Third, while assisting the operation of current grid topologies is valuable, we suggest utilizing GML also for grid expansion and upgrade planning, complementing existing simulation approaches. Distribution grids deserve particular attention here, as operators face mounting operational challenges~\cite{cleenwerckDistributionNetworksTransition2026} yet synthetic dataset availability remains limited compared to transmission grids, a gap that the community should actively work to close.

Fourth, storage optimization represents a growing practical need that GNN research has addressed only partially. As storage technologies scale, the joint optimization of placement, sizing, real-time control strategies, and integration into market operations becomes increasingly important, yet few studies address this chain end-to-end. Graph-structured approaches are well-positioned to tackle the spatial aspects of placement and the coordination of distributed assets.

\paragraph{Common GML challenges.}
GML faces some common challenges regardless of its application domain. 
Overfitting and limited generalization are not unique to graph-based models, 
and in power systems this risk is amplified, as sensor data are often missing 
or incomplete~\cite{kuppannagariSpatioTemporalMissingData2021}, labels may 
exist only for rare events or operating 
regimes~\cite{zhangAdaptiveIdentificationPower2024}, and access can be 
restricted by confidentiality or a lack of standardization. The oversmoothing 
and oversquashing failure modes introduced in Section~\ref{s:graph_machine_learning} 
are particularly acute in power systems: Capturing long-range dependencies at 
the transmission level may require up to 13 message-passing 
layers~\cite{ringsquandlPowerRelationalInductive2021}, yet increasing depth 
simultaneously worsens both phenomena~\cite{akanshaOversquashingGraphNeural2025}, 
creating a direct tension between the required receptive field and model 
expressivity. Expressivity limitations more broadly, such as those revealed 
by the Weisfeiler-Lehman isomorphism 
test~\cite{zhangExpressivePowerGraph2025}, have not been addressed in power 
system applications. This raises an open question: How expressive must GML 
methods be to meet the requirements of power system applications?

Overcoming these issues requires GML architectures and training strategies that control overfitting, overcome oversmoothing and oversquashing, and clarify the expressivity needed for reliable power system applications.

\paragraph{Power system graph modeling.}
In Section~\ref{s:power_grid_is_a_graph} we motivated power grids as  graphs and discussed the range of abstraction levels practitioners work  with. Here, we focus on the challenges that arise once that modeling choice has been made.

A first dimension is scale. At the transmission level, graph sizes are moderate ($10^3$--$10^4$ nodes), small compared to citation or social networks, which are often studied with GML, yet scalability remains a recognized issue for grids spanning countries or continents even at higher levels of abstraction. At the  other extreme, applications such as wind-park forecasting or PV-plant  output prediction operate on comparatively small graphs (of about $10^1$--$10^2$ nodes), where scalability is not the primary bottleneck. Importantly, graph size alone is not decisive: Even small graphs can be challenging due to the density  of physics-induced features and the complexity of the underlying tasks.

A second dimension is topology change. Real-world grids are not static: Lines are switched in and out, faults alter connectivity, and operators reconfigure the network in response to changing conditions. A 
particularly challenging instance arises at the substation level, where node breakers and busbar couplers allow the internal connectivity of a substation to be reconfigured in real time. Unlike line switching, which changes connectivity between substations, busbar reconfiguration changes the topology \emph{within} a substation, creating a combinatorially large space of possible configurations that classical methods struggle to represent and optimize over. GML approaches are a natural candidate here, as the resulting configuration changes map directly onto graph structure~\cite{meng2025flow}.

A third dimension is temporal dynamics. In dynamic graphs, the topology, node and edge attributes evolve over time, and this evolution carries essential information in real-world power system applications. Common solutions use hybrid spatiotemporal GNNs, where graph convolutions encode spatially structured interactions while RNNs or CNNs model temporal 
dependencies~\cite{wuComprehensiveSurveyGraph2021}. An alternative is to formulate message passing as a graph dynamical system using ODEs or PDEs. For example, \citet{karimiSpatiotemporalGraphNeural2021} propose a spatiotemporal GNN for interactions among photovoltaic plants, combining spatial connectivity with temporal dynamics to improve power forecasting. This temporal aspect is tightly linked to real-time and online operation: 
Many critical grid functions depend on timely, high-quality data~\cite{kuppannagariSpatioTemporalMissingData2021}, and increasing renewable penetration raises both system variability and decision speed requirements. Offline models are inadequate during fast phenomena such as cascading failures~\cite{varbellaPowerGraphPowerGrid2024a}. Real-time-capable architectures are therefore necessary for predictive and responsive grid operation.

Addressing these three dimensions together requires GML approaches that 
scale to both large and feature-rich small graphs, handle static and 
dynamic topology changes, incorporate temporal dynamics, and support 
real-time operation.

\begin{figure}
    \centering
    \includegraphics[width=\linewidth]{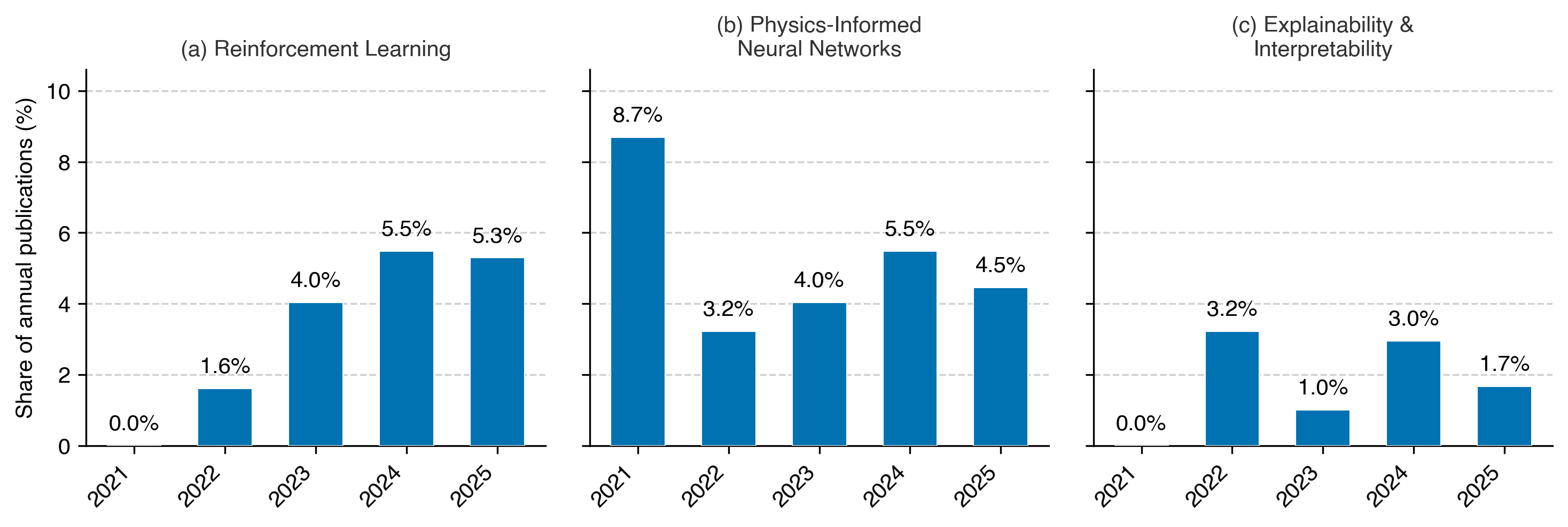}
    \caption{Publication trends for reinforcement learning (RL), physics-informed neural networks (PINN), and explainable AI (XAI) with GML for power systems.}
    \label{fig:gnn_trends}
\end{figure}

\paragraph{Explainability.}
Explainable AI (XAI) is essential for understanding model predictions and for building trust in AI used in critical infrastructure such as power systems. It also supports debugging and scientific discovery~\cite{agarwalEvaluatingExplainabilityGraph2023}. GNNs present unique explainability challenges because their predictions emerge from complex message-passing operations over graph structures, making it difficult to trace which nodes, edges, or subgraphs drove a given output. Standard perturbation-based approaches from non-graph ML do not translate cleanly to the discrete, relational nature of graphs, and evaluation metrics for graph explanations remain largely underdeveloped ~\cite{agarwalEvaluatingExplainabilityGraph2023}. In power system applications, current approaches include intrinsic explainability via physics-informed components ~\cite{pagnierPhysicsInformedGraphicalNeural2021}, post hoc attribution such as subgraph extraction, gradient-based methods, and attention analyses ~\cite{sadricAssessingExplanationsGraph2026a, raumExplainableAIAnalyzing2025}, and visualization-based analysis ~\cite{solheimVisualizingGraphNeural2024}. Counterfactual reasoning (asking which minimal graph perturbation would change a prediction) is a particularly underexplored direction for graph-structured data~\cite{sadricAssessingExplanationsGraph2026a} but would be a great addition, considering the usage of GNNs for grid planning. Despite this progress, explainability remains underused, as shown in Figure ~\ref{fig:gnn_trends}, the annual share of GNN-related publications addressing it has never exceeded roughly three percent. A productive path forward is to combine post hoc techniques with physics-informed models and to develop intrinsically interpretable architectures. Ultimately, explanations must be validated by domain experts through human-in-the-loop evaluation to confirm that they are actionable and meaningful in the context of grid operations.

Advancing explainability in GML therefore requires methods that integrate physics-informed modeling with post hoc techniques, deliver actionable insights, and build the trust needed for deployment in power systems. Physics-informed approaches showed an early peak of nearly nine percent in 2021 before settling in the four to six percent range, suggesting that PINN integration has transitioned from a novel direction into an established modeling component.

%Figure~\ref{fig:gnn_trends} also documents trends for the other two directions covered in this review. Graph-based reinforcement learning (Section~\ref{ss:optimization_and_control}) was virtually absent in 2021 but grew steadily to roughly five percent of annual GML publications by 2024--2025, reflecting rising interest in adaptive, topology-aware control policies. 

Collectively, the challenges discussed in this section point to a common requirement: GML methods for power systems must move beyond proof-of-concept demonstrations toward architectures that are scalable, physically grounded, interpretable, and rigorously evaluated on realistic benchmarks. Addressing these gaps will require sustained collaboration between the GML and power systems communities, shared infrastructure for benchmarking and reproducibility, and a willingness to treat domain constraints not as obstacles but as inductive biases that can improve generalization. The field is well-positioned to make this transition, and the rapid growth in publications documented in Section~\ref{s:applications} suggests that the necessary community momentum is already building.

% 6. Conclusion 

% Goal: A strong summary statement.
% Summarize the potential: GML is not just an efficiency hack; it is likely the enabling technology for a 100% renewable, decentralized grid.

\section{Conclusion}
\label{s:conclusion}
% no outlook as we already have a lot of ideas for future research in previous sections 4, 5, and 6. Here we don't repeat ourselfs just recall why this review is important and why it was written as well put it into broader scope and end with motivation

Graph-based machine learning has progressed beyond the proof-of-concept stage in power systems. It now supports real-world applications, such as forecasting, monitoring, and decision-making. From an engineering standpoint, GML provides new abstractions that complement traditional methods. From an ML perspective, power systems provide a high-impact testing ground for developing robust, generalizable, and inductive models under different homophily levels and strong relational structures. Despite its promise, GML in power systems still faces challenges, including scalability, interpretability, and limitations in data availability and quality.

\paragraph{Foundation models.} The breadth of these applications (forecasting, grid analysis and state estimation, optimization and control, fault diagnosis and reliability, and cybersecurity) naturally raises the question: Why not unify them under a single model? ~\citet{hamannFoundationModelsElectric2024} recently proposed GridFM, a foundation model specifically designed for the electric power grid.  It is pre-trained on large 
volumes of grid data using self-supervised learning to capture underlying patterns. This allows the model to be adapted and fine-tuned efficiently for many power system applications. It offers both flexibility and significant computational speed-ups. According to the authors, GridFM's idea is to handle complexity and uncertainty more effectively than conventional methods and should provide a platform that enables stakeholders to fine-tune the model with their own proprietary data in a scalable and cost-effective way. So far, the initial version GridFM-v0 is limited to power flow applications and further extensions are planned for the next years. It will be interesting to monitor whether a single foundation model, such as GridFM or other approaches, such as the work by~\citet{tuPowerPMFoundationModel2024a}, will address all graph-based challenges for power systems or whether a collection of specialized models will remain in use.

\paragraph{Relation to open challenges in GML.} Our review engages directly with open questions raised in recent literature. In their position paper, ~\citet{bechler-speicherPositionGraphLearning2025} identify four major limitations in GML research: The absence of transformative real-world applications, poorly constructed graphs, weak benchmarking culture, and the lack of foundational models. We demonstrate that power systems provide evidence against the first two limitations: Applications on real-world data already exist, and power grid graphs are naturally and meaningfully structured. However, we share concerns about benchmarking and foundational modeling, which remain open issues. Progress on these open issues will hinge on closer collaboration between researchers in power systems and machine learning. 

\vspace{1em}
GML and power systems are converging in ways that are both timely and essential. ~\citet{rolnickTacklingClimateChange2023} highlight that enabling low-carbon electricity systems is among the highest-impact applications of machine learning in addressing climate change. The applications surveyed in this review, from renewable generation forecasting to optimal power flow and grid resilience, directly contribute to this goal. Despite the impressive breadth of applications documented here, many results remain difficult to verify or extend: Models are rarely shared, datasets are often inaccessible or insufficiently documented, and simulation protocols go unreported. We therefore call on the community to prioritize dedicated benchmark studies and the release of open datasets and models. Closing these reproducibility gaps is essential for the scientific credibility of the field, and will, alongside sustained collaboration between the GML and power systems communities, determine how quickly these advances reach operational deployment.

\appendix
\section{Literature Search Methodology}
A systematic literature search was conducted using Scopus and the Scopus beta preprint search to identify relevant publications on the application of Graph Neural Networks to power grids and power systems. The search was performed using the query \texttt{TITLE-ABS-KEY("Graph Neural Networks" AND ("Power Grid" OR "Power Systems"))} across title, abstract, and keywords fields. To ensure comprehensive coverage of both peer-reviewed and early-stage research, the Scopus preprint search (beta) was also consulted, drawing from indexed preprint servers including arXiv, bioRxiv, ChemRxiv, medRxiv, Research Square, SSRN, and TechRxiv. The search was limited to publications up to and including December 31, 2025. In addition to the structured database search, the authors incorporated further relevant works known from their own prior research experience. Each identified paper was then manually reviewed, at minimum by title and abstract, to assess its relevance, with papers deemed out of scope excluded and the remaining qualifying papers subsequently classified into one of the defined application areas.

\section*{Declaration of Generative AI and AI-Assisted Technologies 
in the Manuscript Preparation Process}
During the preparation of this work, the authors used Claude (Anthropic) 
to support the improvement of language and readability, structuring of 
sections, and refinement of the abstract, as well as to generate code 
for data visualizations (Figures~3--6), which were subsequently verified 
and finalized by the authors. Gemini (Google) was used solely to generate 
code for the companion webpage. No AI-assisted 
tools were used in the literature search or synthesis process. 
After using these tools, the authors reviewed and edited all content 
as needed and take full responsibility for the content of the 
published article.

\printcredits

%\bibliographystyle{cas-model2-names}
\bibliographystyle{unsrtnat}

\bibliography{bibliography/references_elsevier}

\end{document}